\documentclass[12pt,a4paper]{article}
\usepackage[round]{natbib}
\usepackage[T1]{fontenc}
\usepackage{multirow}
\usepackage{pifont}
\usepackage{setspace}
\usepackage{rotating} 
\usepackage{chngcntr}
\counterwithin{figure}{section}
\usepackage{tikz}
\usetikzlibrary{arrows,shapes}
\usetikzlibrary{calc}
\usepackage{graphicx}
\usepackage{framed}

\usepackage{algorithm,algpseudocode}
\makeatletter
\renewcommand{\fnum@algorithm}{\fname@algorithm}
\makeatother

\graphicspath{{./images/}}
\usepackage[numbered]{bookmark}

\usepackage{amsfonts,enumerate,amsmath}
\usepackage{latexsym}
\usepackage{color}

\usepackage{verbatim}
\usepackage{graphicx}
\usepackage{array}
\usepackage{float,bm}
\usepackage{multirow}

\usepackage{appendix}

\graphicspath{{./images/}}

\newcommand{\bqn}{\begin{eqnarray*}}
\newcommand{\eqn}{\end{eqnarray*}}

\DeclareMathOperator{\var}{var}
         
      \DeclareMathOperator{\diag}{diag}

\numberwithin{equation}{section}  

\newtheorem{thm}{Theorem}[section]

\newtheorem{rem}{Remark}[section]
\newtheorem{asump}{Assumption}[section]

\newtheorem{innercustomasump}{Assumption}
\newenvironment{customasump}[1]
  {\renewcommand\theinnercustomasump{#1}\innercustomasump}
  {\endinnercustomasump}

\newtheorem{prp}{Proposition}[section]
\hypersetup{
  colorlinks   = true, 
  urlcolor     = blue, 
  linkcolor    = blue, 
  citecolor   = blue 
}
\newcommand{\blind}{1}

\begin{document}

\def\spacingset#1{\renewcommand{\baselinestretch}%
{#1}\small\normalsize} \spacingset{1}

\spacingset{1.2}


\if1\blind
{
\title{\bf Statistical Learning for Individualized Asset Allocation} 

\author{Yi Ding\thanks{Assistant Professor, Faculty of Business Administration, University of Macau, Macau (e-mail: ydingust@gmail.com)
}, Yingying Li\thanks{Corresponding author.  Professor, Department of ISOM and Department of Finance, Hong Kong University of Science and Technology, Clear Water Bay, Kowloon, Hong Kong (e-mail:yyli@ust.hk)}, \, and  Rui Song 
\thanks{Corresponding author. Professor, Department of Statistics, North Carolina State University, Raleigh, North Carolina, USA (e-mail: rsong@ncsu.edu).   }}
 \date{October 12, 2022}

\maketitle} \fi

\if0\blind
{
  \bigskip
  \bigskip
  \bigskip
  \begin{center}
    {\Large \bf Statistical Learning for Individualized Asset Allocation}
\end{center}
  \medskip
} \fi

\bigskip
\begin{abstract}
\noindent
We establish a high-dimensional statistical learning framework for individualized asset allocation. Our proposed methodology addresses continuous-action decision-making with a large number of characteristics. We develop a discretization approach to model the effect of continuous actions and allow  the discretization frequency to be large and diverge with the number of observations. We estimate the value function of continuous-action using penalized regression with our proposed generalized penalties that are imposed on  linear transformations of the model coefficients. We show that our proposed Discretization and Regression with generalized fOlded concaVe penalty on Effect discontinuity (DROVE)  approach enjoys desirable theoretical properties and allows for statistical inference of the optimal value associated with optimal decision-making. Empirically, the proposed framework is exercised with the Health and Retirement Study data in finding individualized optimal asset allocation. The results show that our   individualized optimal strategy improves the  financial well-being of the population.  
\end{abstract}


\newpage

\spacingset{1.5} 

\section{Introduction}

With the rapid development of artificial intelligence,  particularly  machine learning, a revolution is underway in individualization. In precision medicine, thousands or millions of genetic characteristics are taken into consideration to determine the optimal treatment for an individual patient. Large retail corporations analyze massive datasets of the behavior and personal characteristics of customers to tailor their offerings to individual customers.  

Individualization is not new in asset allocation. For example, before providing financial advice, investment companies ask their customers to  answer questionnaires in order to determine their  customers' risk profiles. An asset allocation strategy would then be recommended in the form of, for instance, a given proportion of stocks versus  bonds that the customer should hold in their portfolio. However, often in these questionnaires, many of the questions are subjective and/or hypothetical. Incorporating 
objective data of the investors can ameliorate the weakness of the reliance of such personalization on subjective opinion.

In this study, we develop a statistical learning framework for individualized asset allocation. We focus on finding an individualized optimal proportion of wealth that should be invested in stocks for a  consumption-based utility optimization problem.  
The essence of the problem  is decision-making, where the action can take value on a continuous set. More broadly, our study provides a theoretical basis and practical showcase for continuous-action decision-making research, such as dose decision in precision medicine,   movement angle in robotic control, and  campaign duration in  personalized marketing.

\subsection*{Related Literature}

Methodologically, our study is closely related to the studies seeking  the optimal individualized treatment rule (ITR) in fields such as personalized medicine. Q-learning and A-learning are  the most popular methods for finding the optimal ITR. Q-learning (\citealt{watkins1989learning}) models the treatment responses, and A-learning (\citealt{murphy2003optimal,robins2004optimal}) models the contrast (regret) function; see, for example, \cite{qian2011performance, zhao2012estimating, Shi2017, athey2019efficient, ZZS19}.  These studies focus primarily on the problem in which the treatment comes from a fixed number of discrete levels, typically binary treatment. 
 
 There is a growing interest in the study of continuous-action decision-making  where the treatment comes from a continuous set. \cite{laber2015tree}  propose a direct tree-based optimal rule searching method. \cite{chen2016personalized} propose a direct nonparametric method by extending the outcome-weighted learning method of \cite{zhao2012estimating}. More recently, \cite{cai2020deep} propose an off-policy evaluation method by adaptively discretizing the action space using a deep jump Q-learning;  \cite{zhu2020kernel} study a kernel assisted optimal dose rule method; while \cite{zhouparsimonious} propose a dimension reduced kernel approximation method. These studies focus on the estimation of optimal policy or  the effect of continuous policy with low-dimensional covariates.

Theoretically,  our study relates to the literature on penalized regression,  especially in high-dimensional settings. The well-known approaches of penalized regression include the least absolute shrinkage and selection operator (LASSO, \citealt{tibshirani1996regression}), smoothly clipped absolute deviation  (SCAD, \citealt{fan2001variable}), and minimax concave penalty (MCP, \citealt{zhang2010nearly}).  The theoretical properties and implementation of the penalized regression estimator under high-dimensional settings are investigated by  \cite{meinshausen2006high}, \cite{zhao2006model}, \cite{zhang2008sparsity}, \cite{lv2009unified}, \cite{fan2011nonconcave}, \cite{wang2013calibrating}, among others. These studies investigate the problem with penalties imposed directly on model coefficients. For the generalized penalties that are imposed on a linear transformation of coefficients, \cite{tibshirani2011solution} and  \cite{arnold2016efficient} investigate the generalized penalty problems using lasso penalties (generalized lasso) and their studies focus mainly on the computational aspect. \cite{she2010sparse} discusses the sparsity recovery property of the generalized lasso estimator when the number of variables is fixed.

In economics and finance, our study relates to the literature on household finance for the elderly and studies that use Health and Retirement Study (HRS)  and Consumption and Activities Mail Survey (CAMS) data.  Using HRS and CAMS data, the  ``retirement savings puzzle'' (\citealt{banks1998there,palumbo1999uncertain}) is studied by \cite{hurd2003retirement} and  \cite{haider2007there}.  \cite{rosen2004portfolio},  \cite{hong2004social} and \cite{de2014bequests} study the problem of asset allocation modeling. By analyzing HRS and~CAMS data, \cite{engen1999adequacy} and \cite{munnell2012national}  underscore a universal inadequacy of post-retirement savings.  
\subsection*{Main Challenges and Our Contributions}

Developing methodologies for continuous-action decision making is challenging. 
Today, it is common to collect a large number of characteristics and incorporate them into the decision-making process. When there are many action levels,  the dimensionality of the problem becomes even larger, posing challenges in identifying  informative characteristics and estimating the model. The policy value, especially the optimal value, is an important target of inference for policy evaluation. For continuous-action decision-making, there are infinite possible options in the search for the optimal policy, posing substantial challenges to the statistical inference of the optimal value.

Empirically, substantial challenges come from the data. The HRS and CAMS survey data contain noisy and incomplete observations. The observed stock ratio data are skewed and concentrated around zero. In addition, the important utility variable is unobserved. Data preparation needs to be carefully done in order to obtain sufficient high-quality data for the statistical learning.

The main contributions of our paper are as follows. 

First, in the field of individualized decision-making, we develop a high-dimensional statistical learning framework to study continuous-action decision-making. Specifically, we develop a discretization approach to model the effect of continuous actions and allow  the discretization level to be large and diverge with the number of observations. The size of the discretization level is carefully analyzed in order to balance  the approximation accuracy of the discretization with the compatibility in the penalized regression framework. 

Second, we propose a Discretization and Regression with generalized fOlded concaVe penalty on Effect discontinuity (DROVE) method to estimate the model  for continuous-action decision-making, whose design accommodates the continuous nature of the actions. Different from the standard penalized regression that penalizes the coefficients directly,~DROVE penalizes the effect differences between adjacent action levels.   Our estimator enjoys model coefficient estimation consistency.

Third, our novel approach provides valid statistical inference for the optimal value of continuous action when there are a large number of characteristics. We obtain the central limit theorem for the proposed DROVE estimator of the model coefficients as well as the value that is associated with  optimal decision-making.
To the best of our knowledge, our study  is the first to achieve  optimal value inference for continuous action, especially under a setting with high-dimensional characteristics. 

Last but not least, in the fields of economics and finance, as a pioneer work, our study provides an individualized asset allocation using a high-dimensional statistical learning method that processes personal characteristic information. To address the data challenges, we use trajectory path models, which allow us to generate pseudo consumption and income for randomly assigned stock ratios. We then obtain the utility  from the pseudo consumption paths and individual risk aversion models. 
Our empirical exercise with HRS and CAMS data shows that our individualized optimal asset allocation strategy substantially improves the financial well-being of the population  and surpasses benchmark strategies that assign fixed stock ratios to all households. The superior performance of our method  demonstrates the importance of individualization in asset allocation.  
\vskip 0.4cm

The paper proceeds as follows. We present the statistical learning framework in Section~\ref{pi_frame} and develop the theoretical properties in Section~\ref{Math}. Sections~\ref{Simu} and~\ref{Empi} are devoted to simulation and empirical studies, respectively. Section~\ref{Conc} contains concluding remarks.

\section{Statistical Learning Framework for Continuous-Action Decision-Making}\label{pi_frame}
 \subsection{Model Setup}

Suppose that we have $n$ observations~$(\mathbf{X}_i,A_i, Y_i)_{i=1, ..., n}$, where~$\mathbf{X}_i\in \boldsymbol{\mathcal{X}}$ is a length-$d$ vector of covariates, $A_i$ is the action that comes from continuous support $\mathcal{A}$, and~$Y_i$ is a random outcome. Under the potential outcome framework (\citealt{rubin1974estimating}), we denote $Y^a$ as the potential outcome that would have been observed under action level $a\in \mathcal{A}$. Following the literature of causal inference (e.g., \citealt{robins2004optimal}), we make the stable unit treatment value assumption (SUTVA) that $Y=Y^a$ if the action $A=a$ (consistency). In addition, we consider the randomization assumption that $A$ is independent of $\mathbf{X}$ and the potential outcomes (e.g., \citealt{murphy2005experimental}), and the positivity assumption on the  density function of $A\in\mathcal{A}$ (e.g., \citealt{chen2016personalized}).  
A deterministic policy $\pi$ maps the features space to the action space, $\pi:\boldsymbol{\mathcal{X}}\to\mathcal{A}$. Under the SUTVA assumption, the optimal policy $\pi^*$ determines the optimal action that maximizes the expected reward given the characteristics~$\mathbf{X}$, that is, 
 \mbox{$\pi^*(x)=\mathop{\arg\max}_a E(Y^a|\mathbf{X}=x, A=a)=\mathop{\arg\max}_a E(Y|\mathbf{X}=x, A=a).
$}

Under the above framework, a particular example of interest  is finding the optimal individualized asset allocation.  For individualized asset allocation,  $\mathbf{X}$ is individual characteristics, such as the financial and demographic status; $Y$ is the economic reward or  utility;~$A$ is the proportion of total wealth invested in stocks, that is, stock ratio,  which can be continuous between 0 and 1; and~$\pi^*(\cdot): \mathbb{R}^d\to [0,1]$ is the investment decision rule that yields the optimal stock ratio given the individual characteristics.

We study the optimal continuous-action decision-making by modeling the conditional expected reward  as a value function $Q(x, a)= E(Y|\mathbf{X}=x, A=a)$. 
We consider the following model
\begin{equation}\label{Q_learning_presentation}
Y=Q_{n}(\mathbf{X}, A)+\varepsilon, \quad\quad Q_n(\mathbf{X}, A)=\boldsymbol{\psi}_0^T \mathbf{X}+\sum_{k=2}^{L_n} \boldsymbol{\psi}^T_{k}\mathbf{X}\cdot\mathbf{1}_{\{A\in [A_{(k)},A_{(k+1)})\}},
\end{equation}
where $Q_n(\mathbf{X}, A)$ is the value function and  $\varepsilon$  is the noise term that is independent of $\mathbf{X}$, 
 $\{0=A_{(1)}< ...<A_{(L_n)}=1\}$ is a series of grids, ~$L_n\leq n$ is the number of grids that can be large and increase with sample size $n$, and $\boldsymbol{\psi}_0, \boldsymbol{\psi}_2, ..., \boldsymbol{\psi}_{L_n}$, are $d$-dimensional vectors,~$\boldsymbol{\psi}_1=\mathbf{0}$ and~\hbox{$A_{(L_n+1)}>1$}.
The two components, $\boldsymbol{\psi}_0^T \mathbf{X}$ and $\sum_{k=2}^{L_n} \boldsymbol{\psi}^T_{k}\mathbf{X}\cdot\mathbf{1}_{\{A\in [A_{(k)},A_{(k+1)})\}},$ represent the main effect of characteristics and the {treatment effect of the action on individual characteristics}, respectively.

We write $\check{\mathbf{X}}_i=\big((\mathbf{X}_i)^T, (\mathbf{X}_i)^T\cdot\mathbf{1}_{\{A_i\in [A_{(2)},A_{(3)})\}}, ..., (\mathbf{X}_i)^T\cdot\mathbf{1}_{\{A_i\in [A_{(L_n)},A_{(L_n+1)})\}}\big)^T,$ and $\boldsymbol{\beta}_n=(\beta_1, ..., \beta_p)^T=(\boldsymbol{\psi}_0^T, \boldsymbol{\psi}^T_{2}, ..., \boldsymbol{\psi}^T_{L_n})^T$  as a length-$p$ vector of the coefficients, 
where $p=d\times L_n$. Hence, the value function can be represented as $Q_n(\mathbf{X}_i, A_i)=\boldsymbol{\beta}_n^T\check{\mathbf{X}}_i.$
We allow $p$  to be large, such that $\log p=o(n)$. In particular, we include the case where characteristics space can be large, so that $\log d=o(n)$. As to the discretization level $L_n$, it can grow slowly with $n$ so that the approximation error in the working model diminishes to zero as~$n\to\infty$. We discuss the growth rate of $L_n$ in more detail in Section \ref{Penalize_reg}. Under the working model~\eqref{Q_learning_presentation}, we denote the true population parameter as $\boldsymbol{\beta}_n^{\star}=(\boldsymbol{\psi}_0^{\star T}, \boldsymbol{\psi}^{\star T}_{2}, ..., \boldsymbol{\psi}^{\star T}_{L_n})^T$.

\subsection{Penalized Regression with Generalized Penalties}\label{Penalize_reg}
 
In order to estimate the  coefficients $\boldsymbol{\beta}_n^{\star}$ in the high-dimensional Q-function \eqref{Q_learning_presentation}, we develop a novel penalized regression formulation, which accommodates the fact that decisions are made from a continuum. Intuitively, when the discretization level $L_n$ is large, the distance between adjacent action levels $|A_{(k)}-A_{(k+1)}|$ becomes small and so is the difference in their effects, $(|\boldsymbol{\psi}_{k,j}-\boldsymbol{\psi}_{k+1,j}|)_{1\leq j\leq d}$. We impose penalties on~$|\boldsymbol{\psi}_{k,j}-\boldsymbol{\psi}_{k+1, j}|$ to shrink the difference in effect between two adjacent decisions,~$A_{(k)}$ and~$A_{(k+1)}$, on each covariate~\hbox{$j=1, ..., d$}.

Formally, we write the following penalized regression:
\begin{equation}\label{penaltyreg_QL}
\min_{\boldsymbol{\beta}_n=(\boldsymbol{\psi}_0^T , ..., \boldsymbol{\psi}^T_{L_n})^T}\sum_{i=1}^n (Y_i-\boldsymbol{\beta}_n^T\check{\mathbf{X}}_i)^2+n\sum_{k=0}^{L_n}p_{\lambda_n}(\boldsymbol{\psi}_k)+n\sum_{k=2}^{L_n-1}\sum_{j=1}^{d} p_{\lambda_n}(\boldsymbol{\psi}_{k,j}-\boldsymbol{\psi}_{k+1,j}),
\end{equation}
where the penalty function is $p_{\lambda_n}(\boldsymbol{\xi}):=\sum_{i=1}^h p_{\lambda_n}(\xi_i)$ for any \hbox{vector $\boldsymbol\xi=(\xi_1, \xi_2, ..., \xi_h)^T$}, and~$p_{\lambda_n}( \cdot )$ is a penalty function with tuning parameter~$\lambda_n$. The first penalties $\sum_{k=0}^{L_n}p_{\lambda_n}(\boldsymbol{\psi}_k)$ penalize the main effect and the treatment effect, similar to the binary case; see, for example, \cite{ZZS19}. The second penalties $\sum_{k=2}^{L_n-1}\sum_{j=1}^{d} p_{\lambda_n}(\boldsymbol{\psi}_{k,j}-\boldsymbol{\psi}_{k+1,j})$ penalize the discontinuity. 
\begin{rem}

The penalties, $\big(p_{\lambda_n}(\boldsymbol{\psi}_{k,j}-\boldsymbol{\psi}_{k+1,j})\big)_{2\leq k\leq L_n-1, 1\leq j\leq d}$, generalize the idea of fused lasso (\citealt{tibshirani2005sparsity}), which imposes penalties on the difference in adjacent coordinates $|\beta_i-\beta_{i+1}|$.  \cite{she2010sparse}, \cite{tibshirani2011solution} and \cite{arnold2016efficient} study the generalized lasso problem, which is a penalized regression as formulated in~\eqref{generalized_penalty_QL} with the lasso penalty function for $p_{\lambda}(\cdot)$. In contrast to these works, we propose a generalized folded concave penalty and investigate the statistical properties of the estimator under the high-dimensional setting. 
 In the literature on individualized decision making, to the best of our knowledge, it is the first time that regression with a generalized folded concave penalized is formulated in studying the effect of continuous actions.
\end{rem}

The proposed penalized regression \eqref{penaltyreg_QL} can be categorized as one with \emph{generalized} penalties, where  penalization is imposed on linear transformations of the coefficients, $\mathbf{D}\boldsymbol{\beta}_n$, for some $K\times p$ matrix $\mathbf{D}=(\mathbf{d}_1, ..., \mathbf{d}_K)^T$:
\begin{equation}\label{generalized_penalty_QL}
\min_{\boldsymbol{\beta}_n} \frac{1}{n}\sum_{i=1}^n (Y_i-\boldsymbol{\beta}_n^T\check{\mathbf{X}}_i)^2+p_{\lambda_n}(\mathbf{D}\boldsymbol{\beta}_n).
\end{equation}
More generally, we consider the following generalized linear model (\citealt{fan2011nonconcave})   where the density function~$f(\mathbf{Y}; \check{\mathbf{X}}, \boldsymbol{\beta}_n)$ satisfies
\begin{equation}\label{def:f}
f(\mathbf{Y}; \check{\mathbf{X}}, \boldsymbol{\beta}_n)=\prod_{i=1}^n f_0(Y_i; \boldsymbol{\beta}_n^T\check{\mathbf{X}}_{i})=\prod_{i=1}^n \exp \bigg\{\frac{Y_i\boldsymbol{\beta}_n^T\check{\mathbf{X}}_{i}-b(\boldsymbol{\beta}_n^T\check{\mathbf{X}}_{i})}{\phi}\bigg\} c(Y_i, \phi),
\end{equation}
where $\boldsymbol{\beta}_n=(\beta_1, ..., \beta_p)^T$ is a $p\times 1$ vector of regression coefficients, $\phi\in (0, \infty)$ is the nuisance parameter of dispersion,  $\check{\mathbf{X}}=(\check{\mathbf{X}}_{1}, \check{\mathbf{X}}_{2}, ..., \check{\mathbf{X}}_{n})^T$, { $b(\cdot)$  is twice continuously differentiable with~$b''(\cdot)>0$, and $c(Y_i, \phi)$ is the base measure that represents the density function for~$Y_i$ when $\boldsymbol{\beta}_n=\mathbf{0}$, for example, $c(Y_i, \phi)=(\pi \phi)^{-1/2}\exp(-Y_i^2/\phi)$ for normal density}.  For a given  matrix~$\mathbf{D}$, the penalized likelihood function with  generalized penalties is
\begin{equation}\label{def:Q}
\aligned
\mathcal{L}_n(\boldsymbol{\beta}_n, \lambda_n, \mathbf{D})=&l_n(\boldsymbol{\beta}_n)-\sum_{k=1}^K p_{\lambda_n}(\mathbf{d}_k^T\boldsymbol{\beta}_n)=\frac{1}{n}\big(\mathbf{Y}^T\check{\mathbf{X}}\boldsymbol{\beta}_n-\mathbf{1}^T\mathbf{b}(\check{\mathbf{X}}\boldsymbol{\beta}_n)\big)-\sum_{k=1}^K p_{\lambda_n}(\mathbf{d}_k^T\boldsymbol{\beta}_n),
\endaligned
\end{equation}
where $l_n(\boldsymbol{\beta}_n)=\big(\mathbf{Y}^T\check{\mathbf{X}}\boldsymbol{\beta}_n-\mathbf{1}^T\mathbf{b}(\check{\mathbf{X}}\boldsymbol{\beta}_n)\big)/n$  and $\mathbf{b}(\check{\mathbf{X}}\boldsymbol{\beta}_n)=\big(b(\boldsymbol{\beta}_n^T\check{\mathbf{X}}_{1}), ..., b(\boldsymbol{\beta}_n^T\check{\mathbf{X}}_{n})\big)^T$. 

We next write $\mathbf{D}=(\mathbf{D}_{signal}^T, \mathbf{D}_{null}^T)^T$, where  $\mathbf{D}_{signal}$ is a~$K_1\times p$ matrix,  $\mathbf{D}_{null}$ is a $K_0\times p$ matrix, and $K=K_0+K_1$. 
Suppose that the true coefficients,~$\boldsymbol{\beta}_n^{\star}=(\beta^{\star}_{1}, ..., \beta^{\star}_{p})^T$, satisfy $\mathbf{D}_{null}\boldsymbol{\beta}_n^{\star}=\mathbf{0}$, that is, $\mathbf{d}_k^T\boldsymbol{\beta}_n^{\star}=0$, for $k=K_1+1, K_1+2, ..., K$. 
In terms of $\mathbf{D}_{signal}$, we impose no constraint on its shape or rank. 

For any  positive semi-definite matrix \hbox{$\mathbf{A}=(a_{ij})$},  we define $\|\mathbf{A}\|_2=\max_{\|\mathbf{x}\|_2\leq 1}\|\mathbf{A}\mathbf{x}\|_2$ and $\|\mathbf{x}\|_2=\sqrt{\sum x_i^2}$ for any vector $\mathbf{x}=(x_i)$. 

We impose the following assumptions on $\mathbf{D}$ and $\rho (t, \lambda)=p_\lambda(t)/\lambda$.
\begin{asump}\upshape\label{B_i}
$\zeta_{\min}^+(\mathbf{D}_{null}\mathbf{D}_{null}^T)\geq c$ and
$\max_{1\leq k\leq K}\|\mathbf{d}_{k}\|_2\leq C$ for some constants $c, C>0$, 
where $\zeta_{\min}^+(\mathbf{D}_{null}\mathbf{D}_{null}^T)$ denotes the smallest nonzero eigenvalue of  $\mathbf{D}_{null}\mathbf{D}_{null}^T$.  
\end{asump}
\begin{rem}Assumption \ref{B_i} is met by our design of $\mathbf{D}$ in \eqref{penaltyreg_QL} as $\zeta_{\min}^+(\mathbf{D}_{null}\mathbf{D}_{null}^T)\geq 1$ and $\max_{1\leq k\leq K}\|\mathbf{d}_k\|_2\leq \sqrt{2}$.  
\end{rem}

\begin{asump}\upshape\label{B_ii}
(1) $\rho (t, \lambda)$ is increasing and concave in $t\in[0, \infty)$; (2) $\rho (t, \lambda)$ is differentiable in $t\in(0, \infty)$ with $\rho' (0+, \lambda)>0$; and (3) if $\rho' (t, \lambda)$  is dependent on $\lambda$, $\rho' (t, \lambda)$ is increasing in $\lambda\in(0, \infty)$ and $\rho' (0+)$ is independent of $\lambda$. 
\end{asump}
\begin{rem}
Assumption~\ref{B_ii} describes the characteristics of a folded concave function class; see, for example, \cite{lv2009unified, fan2011nonconcave}.  Popular examples of folded concave functions include SCAD (\citealt{fan2001variable}) and MCP (\citealt{zhang2010nearly}).  Our numerical examples uses SCAD as representative of the folded concave penalty function family.
\end{rem}
We define~$s_n=p-\text{rank}(\mathbf{D}_{null})$. We regard $s_n$ as the nonsparsity coefficient, which is a natural extension of the nonsparsity coefficient for standard penalized regression when $\mathbf{D}$ is an identity matrix. {The  definition of $s_n$ is in line with that of the degrees of freedom for generalized lasso (\citealt{tibshirani2011solution,tibshirani2012degrees}).} The true model is considered to be sparse in the sense that~$s_n\ll n$. The sparsity of the model comes from two sources. One source of sparsity is the large covariate space in which there can be many irrelevant variables. In real applications, for example, investment agencies collect a large number of covariates, although only a small proportion of the variables are useful.  The sparsity that comes from high-dimensional covariates is the usual notion of model sparsity discussed in the literature.
In addition, we assume sparsity on the treatment effect difference between adjacent action levels. This is essentially to assume that the effect of continuous action exhibits smoothness; hence when~$L_n$ is large, many adjacent action levels have roughly the same treatment effect. This is sensible in real applications.  For example, in asset allocation, when there are small changes in stock ratio,
 utility would not vary drastically.

Let  $g_n=2^{-1}\min\{|\mathbf{d}_j^T\boldsymbol{\beta}_n^{\star}|, \mathbf{d}_j^T\boldsymbol{\beta}_n^{\star}\neq 0\}$ be half of the minimum signal. We impose the following assumptions.
\begin{asump}\upshape\label{B_iii}$g_n\gg\lambda_n\gg\max(\sqrt{s_n/n}, \sqrt{(\log p)/n})$, $\max(s_n,\log p)=o(n)$, and 
$p_{\lambda_n}' (g_n)=o\Big(\min(n^{-1/2}s_n^{-1/2}, n^{-1/2}K_1^{-1}s_n^{1/2})\Big)$. 
\end{asump}
Assumption \ref{B_iii} states that the minimal signal should be sufficiently large to be distinguishable from the noise. 
If $g_n$ decreases as  $L_n$ grows, the minimal signal condition would constrain the number of discretization levels that we are able to handle. Asymptotically,~$L_n$ can be $O( n^{\varsigma})$ for some $\varsigma<1/3$. In Appendix~A of the supplementary materials, we discuss the rate of $L_n$ under a varying coefficient example.   We illustrate the choice of $L_n$ with our practical example in Section \ref{simu_setup}. Additional regularity conditions (Assumptions C.1--C.4)  are in Appendix C of the supplementary materials.

{We summarize the proposed approach for continuous-action decision-making: Discretization and Regression with generalized fOlded concaVe penalty on Effect discontinuity (DROVE) as follows.}
\begin{algorithm}[H]
\caption{{DROVE}}
\begin{algorithmic}
\item[Step I.]Discretize the action support into a series of grids $A_{(1)}< ...<A_{(L_n)}$ with  $L_n$ growing with $n$, and $L_n=O(n^{\varsigma})$ for some $\varsigma < 1/3$.
\item[Step II.] Perform the penalized regression with the generalized folded concave penalty on effect discontinuity,  \eqref{penaltyreg_QL}, and obtain the estimated value function~$\widehat{Q}_n(\mathbf{{X}}, A)$.{ The~GLLN algorithm to be introduced in Section \ref{Coef_Oracle} can be used to solve \eqref{penaltyreg_QL}.}
\item[Step III.]The estimated optimal decision making is $\widehat{\pi}^*_{n}(\mathbf{\mathcal{X}}):=\min\{A: \widehat{Q}_n(\mathbf{\mathcal{X}}, A)=\max_{A_{(1)}\leq A\leq A_{(L_n)}}\widehat{Q}_n(\mathbf{\mathcal{X}}, A)\}$ for an individual with characteristics $\mathbf{\mathcal{X}}$, and the optimal value function is
$\widehat{Q}^*_{n}(\mathbf{\mathcal{X}}):=\widehat{\boldsymbol{\psi}}_0^T \mathbf{\mathcal{X}}+\max \Big(0, (\widehat{\boldsymbol{\psi}}^T_{k}\mathbf{\mathcal{X}})_{k=2, ..., L_n}\Big)$. 
\end{algorithmic}
\end{algorithm}

\section{Statistical Properties}\label{Math}

In this section, we present the statistical properties of the proposed DROVE methodology for continuous-action decision-making. 
 The proofs are in Appendix~E of the supplementary materials.

\subsection{Theoretical Properties of Coefficient Estimation}\label{Coef_Oracle}

The following theorems give the statistical properties of our coefficient estimator.

\begin{thm}\label{thm:oracle}
Under Assumptions~\ref{B_i}--\ref{B_iii}, C.1 and C.2, there exists a strict local maximizer $\widehat{\boldsymbol{\beta}}_n=(\widehat{\boldsymbol{\psi}}_0^T, \widehat{\boldsymbol{\psi}}_2^T, ..., \widehat{\boldsymbol{\psi}}_{L_n}^T)^T$ of the penalized likelihood function $\mathcal{L}_n$, which satisfies 
$$
P(\widehat{\boldsymbol{\psi}}_{Z_0}=\mathbf{0})\to 1\, \text{as } n \to \infty,\quad\text{ and }\quad \|\widehat{\boldsymbol{\beta}}_n-\boldsymbol{\beta}_n^{\star}\|_2=O_p(\sqrt{s_n/n}),
$$
where $\widehat{\boldsymbol{\psi}}_{Z_0}=(\widehat{\boldsymbol{\psi}}_{i,j})_{(i,j)\in Z_0} $, $Z_0=\{(i,j):\boldsymbol{\psi}^{\star}_{i,j}=0, i=0, 2, ..., L_n, 1\leq j\leq d\}$. In addition, 
\begin{equation}\label{theta_gfcp}
P\big(\mathbf{D}_{null}\widehat{\boldsymbol{\beta}}_n=\mathbf{0}\big)\to 1\,  \text{as } n \to \infty,\quad\text{ and }\quad \|\widehat{\boldsymbol{\theta}}_n-\boldsymbol{\theta}_n^{\star}\|_2=O_p(\sqrt{s_n/n}),
\end{equation}
where $\widehat{\boldsymbol{\theta}}_n=\mathbf{M}^{-1}\widehat{\boldsymbol{\beta}}_n$ and $\boldsymbol{\theta}_n^{\star}=\mathbf{M}^{-1}\boldsymbol{\beta}^{\star}_n$ for the transformation matrix $\mathbf{M}$ defined in (B.2) in Appendix B.
\end{thm}

\begin{thm}\label{CLT}
Under the assumptions of Theorem \ref{thm:oracle} and Assumption C.3, with probability tending to 1 as $n\to\infty$, the local maximizer in Theorem~\ref{thm:oracle} satisfies
$$
\sqrt{n}\boldsymbol{\Omega}_n (\widehat{\boldsymbol{\beta}}_n-\boldsymbol{\beta}_n^{\star})\xrightarrow[]{\mathcal{D}} \mathcal{N}(0, \phi \mathbf{G}),
$$
where $\boldsymbol{\Omega}_n$ is a $q\times p$ matrix, $q\leq s_n$ and fixed, $n\boldsymbol{\Omega}_n\mathbf{U}_0\mathbf{B}_n^{-1}\mathbf{U}^T_0\boldsymbol{\Omega}_n^T\to \mathbf{G}$, $\mathbf{G}$ is a $q\times q$ positive definite matrix, $\|\mathbf{G}\|_2=O(1)$, $\mathbf{U}_0$ and $\mathbf{B}_n$ are defined in (B.1) in Appendix B and~Assumption~C.3, respectively. 
\end{thm}
In practice, the covariance matrix $\phi \mathbf{G}$  needs to be estimated. Following the conventional technique (e.g., that of \citealt{fan2001variable}), we estimate the variance using the following sandwich formula: $n\boldsymbol{\Omega}_n\mathbf{\widehat{U}}_0\widehat{\mathbf{B}}_n^{-1}\Big(\sum_{i=1}^n(\tilde{\mathbf{ z}}_{i}\tilde{\mathbf{ z}}_{i}^T\widehat{\varepsilon}_i^2)\Big) \widehat{\mathbf{B}}_n^{-1}\mathbf{\widehat{U}}_0^T\boldsymbol{\Omega}_n^T,$ where $\mathbf{\widehat{U}}_0$ is the orthogonal matrix that spans the null space of  $\widehat{\mathbf{D}}_{null}$,  $\widehat{\mathbf{D}}_{null}$ is the sub-matrix of $\mathbf{D}$, which satisfies $\widehat{\mathbf{D}}_{null}\widehat{\boldsymbol{\beta}}_n=\mathbf{0}$, $\widehat{\mathbf{B}}_n=\widetilde{\mathbf{X}}_{signal}^T\mathbf{\Sigma}(\check{\mathbf{X}}\widehat{\boldsymbol{\beta}}_n)\widetilde{\mathbf{X}}_{signal}$,   \hbox{$\widetilde{\mathbf{X}}_{signal}=(\tilde{\mathbf{z}}_{1}, ..., \tilde{\mathbf{z}}_{n})^T=\check{\mathbf{X}}\mathbf{\widehat{U}_0}$}, and $\mathbf{\Sigma}(\boldsymbol{\delta})=\diag\big(b''(\delta_1), ... b''(\delta_n)\big)$ for any $\boldsymbol{\delta}\in \mathbb{R}^{n}$.

According to Theorems~\ref{thm:oracle} and~\ref{CLT}, our coefficient estimator achieves the oracle property in that it identifies the true model with probability tending to one and enjoys strong consistency property. 

Next, with regard to implementation, we introduce the following generalized local linear approximation (GLLA) algorithm, which is a generalization of the LLA algorithm (\citealt{zou2008one, fan2014strong}).
\begin{algorithm}\caption{{Generalized local linear approximation (GLLA)}}
\begin{algorithmic}[Step I]
\item[Step I.]  Initialize $\widehat{\boldsymbol{\beta}}_n^{(0)}=\widehat{\boldsymbol{\beta}}_n^{init}$, and $\mathbf{\widehat{\Theta}}^{(0)}=\big(\widehat{\boldsymbol{\vartheta}}_1^{(0)}, \widehat{\boldsymbol{\vartheta}}_2^{(0)}, ..., \widehat{\boldsymbol{\vartheta}}_K^{(0)}\big)^T$, where $
\widehat{\boldsymbol{\vartheta}}^{(0)}_k=\widehat{w}^{(0)}_k \mathbf{d}_k$, and $\widehat{w}_k^{(0)}=\rho_{\lambda_n}'(|\mathbf{d}_k^{T}\widehat{\boldsymbol{\beta}}_n^{(0)}|)$  for \hbox{$1\leq k\leq K$}.\\
\item[Step II.] For $m$=1, 2, ..., repeat the following till convergence.
\begin{enumerate}[a.]
\item Solve  $\boldsymbol{\widehat{\beta}}_n^{(m)} =\mathop{\arg\max}_{\boldsymbol{\beta}_n} l_n(\boldsymbol{\widehat{\beta}}_n)-\lambda_n \|\mathbf{\widehat{\Theta}}^{(m-1)}\boldsymbol{\widehat{\beta}}_n\|_{1}$, 
where $\|\cdot\|_1$ denotes the $\ell_1$ vector norm, such that $\|\boldsymbol{w}\|_1=\sum_{i=1}^n|w_i|$ for any $\boldsymbol{w}=(w_1,...,w_n)^T$.
\item Update $\mathbf{\widehat{\Theta}}^{(m)}=\big(\widehat{\boldsymbol{\vartheta}}_1^{(m)}, \widehat{\boldsymbol{\vartheta}}_2^{(m)}, ..., \widehat{\boldsymbol{\vartheta}}_K^{(m)}\big)^T$, where~\hbox{$\widehat{\boldsymbol{\vartheta}}^{(m)}_k=\widehat{w}^{(m)}_k \mathbf{d}_k$}, and~{$\widehat{w}_k^{(m)}=\rho_{\lambda_n}'(|\mathbf{d}_k^{T}\boldsymbol{\widehat{\beta}}_n^{(m)}|)$} for  $1\leq k\leq K$. 
\end{enumerate}
\end{algorithmic}
\end{algorithm}
\begin{customasump}{2.2'}\upshape\label{B_iii'}
$\rho' (0+, \lambda)\geq a_1$; $\rho' (t, \lambda)\geq a_1$ for $t\in(0, a_2\lambda)$; $\rho' (t, \lambda)=0$ for $t>a\lambda$ with constants $a > a_2>0$, and $a_1>0$.
\end{customasump}
\begin{rem}
Assumption~\ref{B_iii'} holds for the folded concave penalty, such as SCAD and~MCP.  
\end{rem}

Proposition \ref{LLA} gives the property of the GLLA algorithm. 
\begin{prp}\label{LLA}
Under the assumptions of Theorem~\ref{thm:oracle} and Assumption~\ref{B_iii'}, assume that the oracle estimator $\boldsymbol{\widehat{\beta}}_n$ in Theorem~\ref{thm:oracle} is unique. In addition, assume that
\begin{equation}\label{InitialCond}
\max_{s_n+1\leq j\leq p}|\widehat{\theta}_{nj}^{init}|\leq a_2\lambda_n,\quad and \quad \min_{1\leq k\leq K_1}|\mathbf{d}_k^T\widehat{\boldsymbol\beta}_n^{init}|\geq a\lambda_n,
\end{equation}
for $a_2$ and $a$ defined in Assumption~\ref{B_iii'} and $\boldsymbol{\widehat{\theta}}_{n}^{init}=\mathbf{M}^{-1}\boldsymbol{\widehat{\beta}}_{n}^{init}$.
 Then with probability tending to 1,  the GLLA algorithm initialized by $\widehat{\boldsymbol\beta}_{n}^{init}$ finds the oracle estimator $\widehat{\boldsymbol\beta}_n$ after one iteration. 
\end{prp}

\begin{rem}
One potential choice of the initial value is the generalized lasso estimator, which uses~$\ell_1$ penalty. The algorithm for solving generalized lasso problem is discussed in \cite{tibshirani2011solution} and \cite{arnold2016efficient}.  When $\mathbf{D}=\mathbf{I}$, the error bound exists for standard lasso estimator under proper designs; see (C1) and Corollary 3 in \cite{fan2014strong}. 
\end{rem}

About the computation cost of the algorithm, in particular the computation complexity  with respect to  $L_n$, heuristically, the generalized lasso needs \mbox{$O(\max(K^2 n, K n^2))$} operations; see, for example, \cite{tibshirani2011solution}. The total computation costs of Steps~I and II, therefore, are $O(\max(dL_n,n)dL_nn)$ and $O(\max(dL_n,n)dL_nn)$, respectively. In practice, the maximum iterations can be set as $O(\log n)$ and the algorithm converges fast within a few iterations. The total computation cost of the algorithm is $O\big((\log n)\max(dL_n,n)dL_nn\big)$, which  is polynomial in $L_n$, implying that the algorithm is scalable to high dimensions.

\subsection{Optimal Value Estimation and Inference}\label{OptV_Inf}

One important advantage of our DROVE approach is that it allows for proper inference of the value associated with a decision rule, in particular, the optimal decision  that achieves the maximum value. 

Let $\mathbf{\mathcal{X}}$ be the personal characteristic vector that belongs to the testing population. We use a different notation, $\mathbf{\mathcal{X}}$,  to distinguish the testing population from the estimation sample~$\mathbf{X}$ and stress that $\mathbf{\mathcal{X}}$ is independent of~$\mathbf{Y}$. 
Suppose that the value function \eqref{Q_learning_presentation} holds for~$\mathbf{\mathcal{X}}$; thus it follows that $E(R|\mathbf{\mathcal{X}}, A)=Q_n(\mathbf{\mathcal{X}}, A)$.
Let~$\pi^*_{n}(\cdot)$ denote the optimal decision, and let~$Q^*_{n}(\cdot)$ denote the optimal Q-function associated with~$\pi^*_{n}(\cdot)$. More formally, 
\begin{align}\label{A*:def}
&\pi^*_{n}(\mathbf{\mathcal{X}}):=\min\{A: Q_n(\mathbf{\mathcal{X}}, A)=\max_{A_{(1)}\leq A\leq A_{(L_n)}}Q_n(\mathbf{\mathcal{X}}, A)\},\\
&Q^*_{n}(\mathbf{\mathcal{X}}):={\boldsymbol{\psi}^{\star}_{0}}^T \mathbf{\mathcal{X}}+\max \Big(0, ({\boldsymbol{\psi}^\star_{k}}^T\mathbf{\mathcal{X}})_{k=2, ..., L_n}\Big).\label{Opt_v}
\end{align}
Suppose that $\mathbf{\mathcal{X}}\sim F$, and  the optimal value is $E\big(Q^*_{n}(\mathbf{\mathcal{X}})\big)=\int Q^*_{n}(\mathbf{\mathcal{X}})dF(\mathbf{\mathcal{X}})$. The expectation is taken over the distribution of the testing population $\mathbf{\mathcal{X}}$ given the working model~$Q_n$. More generally, for any given decision rule $\pi(\cdot): \mathbb{R}^d\to [0,1]$ and for~$\mathbf{\mathcal{X}}\in \mathbb{R}^d$,
 we use $\check{\mathbf{\mathcal{X}}}_{\pi}$ to denote a length-$p$ vector
\begin{equation}\label{Def:X_pi}
\check{\mathbf{\mathcal{X}}}_{\pi}=\Big((\mathbf{\mathcal{X}})^T, (\mathbf{\mathcal{X}})^T\cdot\mathbf{1}_{\{\pi(\mathbf{\mathcal{X}})\in [A_{(2)},A_{(3)})\}}, ..., (\mathbf{\mathcal{X}})^T\cdot\mathbf{1}_{\{\pi(\mathbf{\mathcal{X}})\in [A_{(L_n)},A_{(L_n+1)})\}}\Big)^T.
\end{equation} 
Using the notation in \eqref{Def:X_pi}, the optimal Q-function is $Q^*_{n}(\mathbf{\mathcal{X}})={\boldsymbol{\beta}^{\star}_n}^T \check{\mathbf{\mathcal{X}}}_{\pi^*_{n}}$, and the  optimal value is:
$ 
E\big(Q^*_{n}(\mathbf{\mathcal{X}})\big)=\int Q^*_{n}(\mathbf{\mathcal{X}})dF(\mathbf{\mathcal{X}})={\boldsymbol{\beta}^{\star}_n}^T \int \check{\mathbf{\mathcal{X}}}_{\pi^*_{n}}dF(\mathbf{\mathcal{X}})={\boldsymbol{\beta}^{\star}_n}^T E(\check{\mathbf{\mathcal{X}}}_{\pi^*_{n}}).
$

Let $\mathbb{P}_N(\cdot)$ denote the empirical mean measure for a sample of size $N$: $\mathbb{P}_N(\boldsymbol{\omega})=\sum_{i=1}^{N}\omega_i/N$ for any $\boldsymbol{\omega}=(\omega_i)_{1\leq i\leq N}$. Given a testing sample of size $N$,~$(\mathbf{\mathcal{X}}_i)_{1\leq i\leq N}\sim F$, we estimate the optimal value by
\begin{equation}\label{est_opt_value_formula}
\mathbb{P}_N\widehat{Q}^*_{n}(\mathbf{\mathcal{X}})=\frac{1}{N}\sum_{i=1}^N \widehat{Q}^*_{n}(\mathbf{\mathcal{X}}_i)=\widehat{\boldsymbol{\beta}}_{n}^T\Big(\frac{1}{N}\sum_{i=1}^N \check{\mathbf{\mathcal{X}}}_{i,\widehat{\pi}^*_{n}}\Big)=\widehat{\boldsymbol{\beta}}_{n}^T\mathbb{P}_N(\check{\mathbf{\mathcal{X}}}_{\widehat{\pi}^*_{n}}),
\end{equation}
where\begin{align}\label{A*hat:def}
&\widehat{\pi}^*_{n}(\mathbf{\mathcal{X}}):=\min\{A: \widehat{Q}_n(\mathbf{\mathcal{X}}, A)=\max_{A_{(1)}\leq A\leq A_{(L_n)}}\widehat{Q}_n(\mathbf{\mathcal{X}}, A)\},\\
&\widehat{Q}^*_{n}(\mathbf{\mathcal{X}}):=\widehat{\boldsymbol{\psi}}_0^T \mathbf{\mathcal{X}}+\max \Big(0, (\widehat{\boldsymbol{\psi}}^T_{k}\mathbf{\mathcal{X}})_{k=2, ..., L_n}\Big).\label{Opt_est_v}
\end{align}

\begin{rem}
To estimate and make inference about the optimal value, we use the testing sample that is distinct from the Q-function estimation sample.  This approach is similar in spirit to the sample-splitting method in machine learning literature and  is advocated by many recent studies on treatment effect evaluation, such as \cite{chernozhukov2017double}, \cite{wager2018estimation}, and \cite{athey2019efficient}.  The sample-splitting reduces bias and facilitates the valid inference of the optimal value. A discussion about the results of optimal value estimation using the estimation sample is in Appendix D of the supplementary materials. On the other hand, because we are mostly interested in predicting the decision-making effect on a broad population that extends beyond the estimation sample for which the decision-making effects are observable, it is also practically reasonable to use the testing sample to evaluate the effect of optimal decision-making.

\end{rem}

\begin{asump}\upshape\label{B_vii}
Assume 

(1) $s_n=o(n^{\delta_1})$ and $L_n=o(n^{\delta_2})$ for some $\delta_1, \delta_2>0$, and $2\delta_1+\delta_2 <1$.

(2) $(\mathbf{\mathcal{X}}_{j})_{1\leq j\leq N}$ are \hbox{i.i.d.} and independent of $\mathbf{Y}$, $E\big(Q^*_{n}(\mathbf{\mathcal{X}})^2\big)< \infty$, $n=O(N)$, and $N=O(n^M)$ for some $M>1$. There exist some constants $r_0, r_1$ and $r_2>0$, for any $t>0$,   
$P(|\chi_j|> t)\leq r_0\exp(-r_1 t^{r_2})$ for all  $1\leq j\leq d$,
 where $\mathbf{\mathcal{X}}=:(\chi_1, ..., \chi_d)^T$. 
\end{asump}

For the statistical inference of the optimal value, we further impose regularity condition; see Assumption C.4 in Appendix C of the supplmentary materials.

Theorem \ref{CLT_Value} gives the asymptotic distribution of the estimated optimal value.

\begin{thm}\label{CLT_Value}
Under the assumptions of Theorem \ref{thm:oracle} and Assumption \ref{B_vii}, with probability tending to 1 as \hbox{$n, N\to\infty$}, the~$\mathbb{P}_N\widehat{Q}^*_{n}(\mathbf{\mathcal{X}})$ defined in  \eqref{est_opt_value_formula} satisfies

(i) 
\begin{equation}\label{Consist_result_value}
\mathbb{P}_N\widehat{Q}^*_{n}(\mathbf{\mathcal{X}})-E\big(Q^*_{n}(\mathbf{\mathcal{X}})\big)=o_p(1).
\end{equation}

(ii) If, in addition, $2\delta_1+\delta_2<1/2$, Assumptions C.3 and C.4 hold, \\ {$\lim_{n\to\infty}nE{\check{\mathbf{\mathcal{X}}}^T_{\pi^*_{n}}}\mathbf{U}_0\mathbf{B}_n^{-1}\mathbf{U}^T_0E\check{\mathbf{\mathcal{X}}}_{\pi^*_{n}}\phi= \sigma_{1}^2$} and $\lim_{n, N\to\infty}\var\big(Q^*_{n}(\mathbf{\mathcal{X}})\big)n/N= \sigma_{2}^2$ for some constants $\sigma_1>0$, $\sigma_2\geq 0$. Let $\sigma^2_{*}=\sigma_{1}^2+\sigma_{2}^2$, then 
\begin{equation}\label{CLT_result_value}
\sqrt{n}\Big(\mathbb{P}_N\widehat{Q}^*_{n}(\mathbf{\mathcal{X}})-E\big(Q^*_{n}(\mathbf{\mathcal{X}})\big)\Big)\xrightarrow[]{\mathcal{D}} \mathcal{N}(0, \sigma^2_{*}).
\end{equation}
\end{thm}

In practice, in order to apply Theorem~\ref{CLT_Value} to perform feasible statistical inference, we need to estimate the variance. We estimate the variance using the following sandwich formula:
\begin{equation}\label{hat_sig}
\widehat{\sigma}^2_{*}=n\mathbb{P}_N\check{\mathbf{\mathcal{X}}}_{\widehat{\pi}^*_{n}}^T\mathbf{\widehat{U}}_0\widehat{\mathbf{B}}_n^{-1}\Big(\sum_{i=1}^n(\tilde{\mathbf{ z}}_{i}\tilde{\mathbf{ z}}_{i}^T\widehat{\varepsilon}_i^2)\Big) \widehat{\mathbf{B}}_n^{-1}\mathbf{\widehat{U}}_0^T \mathbb{P}_N\check{\mathbf{\mathcal{X}}}_{\widehat{\pi}^*_{n}}+\widehat{\var}\big(\widehat{Q}^*_{n}(\mathbf{\mathcal{X}})\big)n/N,
\end{equation}
where~$\mathbb{P}_N{\check{\mathbf{\mathcal{X}}}}_{\widehat{\pi}^*_{n}}$ and $\widehat{\var}\big(\widehat{Q}^*_{n}(\mathbf{\mathcal{X}})\big)$ are the sample mean of $\check{\mathbf{\mathcal{X}}}_{\widehat{\pi}^*_{n}}$ and sample variance of~$\widehat{Q}^*_{n}(\mathbf{\mathcal{X}})$, respectively, based on the testing sample $(\mathbf{\mathcal{X}}_{j})_{1\leq j\leq N}$. 

\begin{rem}{ 
Our Theorems \ref{thm:oracle} and \ref{CLT_Value} are established for the continuous-action decision making in which the discretization level~$L_n\to\infty$ and hence require intrinsically different methodological design and novel mathematical treatment than existing studies on the binary case (e.g.,  \citealt{shi2016robust}). We device a innovative {DROVE} approach, and obtain a new result \eqref{theta_gfcp} in Theorem \ref{thm:oracle}, which ensures that  the adjacent treatment levels can be identified. Unlike \cite{shi2016robust},  we work under the assumptions for transformed design matrices and transformed parameters (Assumptions C.1--C.3). More essentially,  we consider a more relaxed constraint on the general penalty matrix $\mathbf{D}$. 
 We use a partial reparametrization  technique to show the statistical properties of {DROVE}; see~Appendix~B of the supplmentary materials for more details.  
}
\end{rem}

\begin{rem}For continuous action, there exists no nonparametric/semiparametric approach that yields a $\sqrt{n}$-consistent estimator of effect curve without imposing parametric assumptions; see, e.g., the discussion in \cite{kennedy2017nonparametric}. In this study, on the other hand, we focus on the high-dimensional parametric model \eqref{Q_learning_presentation}, in which the parameter space is sparse in the general sense, hence our approach recovers the true sparse model consistently.  In addition, we allow the discretization level $L_n$ to diverge slowly with $n$, such that $L_n s_n^2=o(\sqrt{n})$, under which setting, inference of the optimal value is obtainable. 
\end{rem}

In addition to the optimal value, our approach also allows for inference of the value difference between the optimal decision and a given decision rule. For a given decision rule, say, $\pi(\cdot):\mathbb{R}^d\to[0,1]$, the  associated Q-function~and the value of~$\pi(\cdot)$ are
$Q_n(\mathbf{\mathcal{X}}, \pi)={\boldsymbol{\psi}^{\star}_0}^T \mathbf{\mathcal{X}}+\sum_{k=2}^{L_n}{\boldsymbol{\psi}^{\star}_k}^T\mathbf{\mathcal{X}}\cdot\mathbf{1}_{\{\pi(\mathbf{\mathcal{X}})\in [A_{(k)},A_{(k+1)})\}}$, 
and
\begin{equation}\label{exp_value_dif}
\aligned
E\big(Q_n(\mathbf{\mathcal{X}}, \pi)\big)&=\int Q_{n}(\mathbf{\mathcal{X}}, \pi)dF(\mathbf{\mathcal{X}})={\boldsymbol{\beta}^{\star}_n}^T E(\check{\mathbf{\mathcal{X}}}_{\pi}).
\endaligned
\end{equation}
The value difference between $\pi^*_{n}$ and $\pi$ is  $E\big(Q^*_{n}(\mathbf{\mathcal{X}})\big)-E\big(Q_n(\mathbf{\mathcal{X}}, \pi)\big)$, which is also called the regret of the policy $\pi$; see, for example, \cite{athey2019efficient}.

We estimate the value associated with $\pi(\cdot)$ using $\mathbb{P}_N\widehat{Q}_{n}(\mathbf{\mathcal{X}}, \pi)$, that is,
\begin{equation}\label{est_value_dif}
\mathbb{P}_N\widehat{Q}_{n}(\mathbf{\mathcal{X}}, \pi)=\frac{1}{N}\sum_{i=1}^N\widehat{Q}_{n}\Big(\mathbf{\mathcal{X}}_i, \pi(\mathbf{\mathcal{X}}_i)\Big)=\widehat{\boldsymbol{\beta}}_{n}^T\mathbb{P}_N(\check{\mathbf{\mathcal{X}}}_{\pi}),
\end{equation}
and we estimate the value difference between  $\pi^*_{n}$ and $\pi$ using $\mathbb{P}_N\widehat{Q}^*_{n}(\mathbf{\mathcal{X}})-\mathbb{P}_N\widehat{Q}_{n}(\mathbf{\mathcal{X}}, \pi)$. 

Proposition \ref{CLT_Value_Dif} gives the asymptotic distribution of the estimated value difference $\mathbb{P}_N\widehat{Q}^*_{n}(\mathbf{\mathcal{X}})-\mathbb{P}_N\widehat{Q}_{n}(\mathbf{\mathcal{X}}, \pi)$. 

\begin{prp}\label{CLT_Value_Dif}
Under the assumptions of Theorem \ref{thm:oracle}  and Assumption \ref{B_vii}, given a decision rule $\pi(\cdot)$: $\mathbb{R}^d\rightarrow [0,1]$, with probability tending to~1 as $n, N\to \infty$, the~$\mathbb{P}_N\widehat{Q}^*_{n}$ and $\mathbb{P}_N\widehat{Q}_{n}(\mathcal{X},\pi)$ defined in  \eqref{est_opt_value_formula} and~\eqref{est_value_dif} satisfy

(i) $\quad
\mathbb{P}_N\widehat{Q}^*_{n}(\mathbf{\mathcal{X}})-\mathbb{P}_N\widehat{Q}_{n}(\mathbf{\mathcal{X}}, \pi)-\Big(EQ^*_{n}(\mathbf{\mathcal{X}})-EQ_n(\mathbf{\mathcal{X}}, \pi)\Big)=o_p(1).
$

(ii) If, in addition, $2\delta_1+\delta_2<1/2$, Assumptions C.3 and C.4 hold,  $\lim_{n\to\infty}nE({\check{\mathbf{\mathcal{X}}}_{\pi^*_{n}}}-{\check{\mathbf{\mathcal{X}}}_{\pi}})^T\mathbf{U}_0\mathbf{B}_n^{-1}\mathbf{U}^T_0E({\check{\mathbf{\mathcal{X}}}_{\pi^*_{n}}}-{\check{\mathbf{\mathcal{X}}}_{\pi}})\phi= \sigma^2_{1}$, and $\lim_{n,N\to \infty}\var\Big(Q^*_{n}(\mathbf{\mathcal{X}})-Q_n(\mathbf{\mathcal{X}}, \pi)\Big)n/N= \sigma^2_{2}$ for some  constants $\sigma_{1}>0$, and $\sigma_{2}\geq 0$. Let $\sigma^2_{*, \pi}=\sigma^2_{1}+\sigma^2_{2}$,  then
$$
\sqrt{n}\Big(\mathbb{P}_N\widehat{Q}^*_{n}(\mathbf{\mathcal{X}})-\mathbb{P}_N\widehat{Q}_{n}(\mathbf{\mathcal{X}}, \pi)-\big(EQ^*_{n}(\mathbf{\mathcal{X}})-EQ_n(\mathbf{\mathcal{X}}, \pi)\big)\Big)\xrightarrow[]{\mathcal{D}} \mathcal{N}(0, \sigma^2_{*, \pi}).
$$
\end{prp}

Similar to \eqref{hat_sig}, we can esimate $\sigma^2_{*, \pi}$ using the following sandwich formula
\begin{multline}\label{hat_sig_2}
\widehat{\sigma}^2_{*, \pi}=n\mathbb{P}_N({\check{\mathbf{\mathcal{X}}}}_{\widehat{\pi}^*_{n}}-\check{\mathbf{\mathcal{X}}}_{\pi})^T\mathbf{\widehat{U}}_0 \widehat{\mathbf{B}}_n^{-1}\Big(\sum_{i=1}^n(\tilde{\mathbf{z}}_{ i}\tilde{{\mathbf{z}}}_{ i}^T\widehat{\varepsilon}_i^2)\Big) \widehat{\mathbf{B}}_n^{-1}\mathbf{\widehat{U}}_0^T\big(\mathbb{P}_N({\check{\mathbf{\mathcal{X}}}}_{\widehat{\pi}^*_{n}}-\check{\mathbf{\mathcal{X}}}_{\pi})\big)\\+\widehat{\var}\big(\widehat{Q}^*_{n}(\mathbf{\mathcal{X}})-\widehat{Q}_n(\mathbf{\mathcal{X}}, \pi)\big)n/N,
\end{multline}
where $\mathbb{P}_N\check{\mathbf{\mathcal{X}}}_{\pi}$ and $\widehat{\var}\big(\widehat{Q}^*_{n}(\mathbf{\mathcal{X}})-\widehat{Q}_n(\mathbf{\mathcal{X}}, \pi)\big)$ are the sample mean of $\check{\mathbf{\mathcal{X}}}_{\pi}$ and sample variance of~$\widehat{Q}^*_{n}(\mathbf{\mathcal{X}})-\widehat{Q}_n(\mathbf{\mathcal{X}}, \pi)$, respectively.

\section{Simulation Study}\label{Simu}
\subsection{Simulation Setup}\label{simu_setup}

We generate data from the model $Y=\boldsymbol{\psi}_0^T\mathbf{X} + \sum_{k=2}^{L_n}\boldsymbol{\psi}_k^T\mathbf{X}\cdot \mathbf{1}_{\{A=A_{(k)}\}}+\varepsilon$ and calibrate the parameters based on the empirical data as described in Section \ref{Empi}. Specifically, the covariates $\mathbf{X}$ is a vector of length~9 and randomly drawn from the personal characteristics that are used in the Q-learning estimation in the empirical study, which includes three binary variables, two categorical variables, three continuous variables and one intercept term.

We divide the continuous action interval $[0,1]$ into 11 discrete levels, i.e.,~\mbox{$L_n=11$}. The stock ratio $A_i$ for each $\mathbf{X}_i$ is randomly assigned from $\{0, 0.1, ..., 1\}$ with equal probability. The choice of $L_n$ is the same as that in the empirical studies. Empirically, in order to balance the applicability of the methodology with the approximation accuracy of the working model, the choice of $L_n$ takes both the theoretical rate and the common practice into consideration. For example, in our personalized asset allocation study using the HRS data, the training sample size is $n=$ 2,000,  $n^{1/3}\approx 13$, and we choose $L_n=11$. On the other hand, it is a common practice for mutual funds to use a~10\% incrementation and make 0\%, 10\%, 20\%, ..., 100\% recommendations.\footnote{See: https://institutional.vanguard.com/assets/pdf/vrpa/InvestorQuestionnaireAssetAllocationInsert.pdf.}   
We illustrate the case when $L_n =20$ in Appendix G of the supplementary materials. When a larger $n$ is available, nice numerical properties can be expected for a larger range of $L_n$.

The Q-function parameters are learned from the empirical study and with hard thresholding (see Appendix F of the supplementary materials for a full description).  The number of coefficients is therefore \mbox{$p=11\times 9=99$}. The noise, $\varepsilon$, is generated from $\mathcal{N}(0, 0.5^2)$.
We design the generalized penalty matrix $\mathbf{D}=(\mathbf{D}_{signal}^T,\mathbf{D}^T_{null})^T$ according to~\eqref{penaltyreg_QL}. 
  The  $\boldsymbol{\beta}_n=(\boldsymbol{\psi}_0^T, \boldsymbol{\psi}_2^T,...,\boldsymbol{\psi}_{L_n}^T)^T$ has~55 zeros, and \mbox{$\text{rank}(\mathbf{D}_{null})=76$}, thus the nonsparsity coefficient is~{$s_n=99-76=23$}. The minimal signal is $g_n=0.4$.

\subsection{Simulation Results}

We estimate the model using the DROVE method.   For comparison, we also evaluate the results using the standard lasso (std-lasso) and standard scad\footnote{{Existing approaches for the decision-making problem under the high-dimensional setting focus on binary problems, among which one of the most comparable  to ours is \cite{shi2016robust}, who use the SCAD estimator to fit Q-function. The std-scad estimator we evaluate, therefore, can be considered as an ad-hoc extension of \cite{shi2016robust} by discretization to the continuous-action  setting. }} (std-scad). We also present the results of the infeasible oracle estimator. The oracle solution is obtained by performing a least squares regression of the response $\mathbf{Y}$ over~$\mathbf{\widetilde X}_{signal}$, as defined in (C.2) in Appendix C. The regression yields the oracle transformed estimator $\widehat{\boldsymbol{\theta}}_n$, and the oracle estimator $\widehat{\boldsymbol{\beta}}_n$  is obtained by the transformation $\widehat{\boldsymbol\beta}_n=\mathbf{M}\widehat{\boldsymbol{\theta}}_n$. The tuning parameters are chosen by minimizing the validation error. 

We conduct 500 independent replications with the sample sizes $n=$ 2000 and~\mbox{$n=$ 3000}. We evaluate the coefficient estimation accuracy of various methods by the $\ell_2$ and $\ell_1$ errors, and measure the parameter selection accuracy by the false positive rate (FP/N, the  number of all false positives divided by that of all total negatives in $\widehat{\boldsymbol\beta}_n$), and false negative rate (FN/P, the number of all false negatives divided by that of all total positives in $\widehat{\boldsymbol\beta}_n$). The total negatives~(N) and the total positives (P) in $\boldsymbol{\beta}_n$ are $55$ and $44$, respectively. {For counting the false positives and false negatives, we set the threshold level to be $10^{-4}$}. The results are in Table~\ref{simu_result_coef_model}. The results show that our {DROVE} estimator has the lowest estimation error  among all compared methods. 
\begin{table}
\spacingset{1}
\caption{{{Coefficient estimation and parameter selection accuracy of various methods }}}
\begin{center}\small
\tabcolsep 0.1in\renewcommand{\arraystretch}{1.1} \doublerulesep
1.2pt
\begin{tabular}{lccccc}
\hline \hline
Error &oracle &{DROVE}  &std-scad&std-lasso  \\\hline
\multicolumn{5}{l}{\;$n=$ 2000}\\
$\|\boldsymbol{\widehat{\beta}}_n-\boldsymbol{\beta}_n\|_{2} $&1.683(0.322)&4.370(0.986)&4.844(0.788)&5.348(0.904)\\
$\|\boldsymbol{\widehat{\beta}}_n-\boldsymbol{\beta}_n\|_{1} $&8.158(1.628)&24.054(7.083)&29.113(6.089)&37.195(6.034)\\
FP/N  &-&0.148(0.083)&0.242(0.109)&0.740(0.082)\\
FN/P &-&0.135(0.070)&0.177(0.069)&0.073(0.046)\\\\
\multicolumn{5}{l}{\;$n=$ 3000}\\
  $\|\boldsymbol{\widehat{\beta}}_n-\boldsymbol{\beta}_n\|_{2} $&1.348(0.239)&3.221(0.862)&3.909(0.723)&4.344(0.699)\\
  $\|\boldsymbol{\widehat{\beta}}_n-\boldsymbol{\beta}_n\|_{1} $&6.558(1.225)&16.486(5.648)&23.033(5.555)&30.466(4.964)\\
  FP/N &-&0.095(0.076)&0.237(0.118)&0.772(0.074)\\
  FN/P &-&0.094(0.068)&0.124(0.062)&0.042(0.035)\\\hline
\end{tabular}
\end{center}\label{simu_result_coef_model}
\small NOTE:  Coefficient estimation accuracy is measured by the $\ell_2$ and $\ell_1$ errors, and parameter selection accuracy is measured by the false positive rate (FP/N) and false negative rate (FN/P). We report the mean and standard deviation~(in parentheses) from~500 replications.
\end{table}

Next, we estimate  the optimal value  and construct its confidence interval (CI). According to Theorem~\ref{CLT_Value}, the confidence interval of the optimal value at a significance level~$\alpha>0$ \\ is $\big(\mathbb{P}_N\widehat{Q}^*_{n}(\mathbf{\mathcal{X}})\big)\pm \widehat{\sigma}(\mathbb{P}_N\widehat{Q}^*_{n})\times \mathcal{N}^{-1}(1-0.5\alpha)$, where $\mathcal{N}(\cdot)$ is the cumulative distribution function of the standard normal distribution, $\widehat{\sigma}(\mathbb{P}_N\widehat{Q}^*_{n})=\widehat{\sigma}^*_{n}/\sqrt{n}$. The feasible variance estimator in \eqref{hat_sig} is used. The estimation sample sizes are $n=$ 2000 and 3000, and the testing sample sizes are $N=$ 5000 and 15,000. The testing sample is randomly generated independently based on empirical data of the personal characteristics. Table~\ref{CI} reports the estimated optimal and empirical coverages of 90\%, 95\% and 99\% CI from 500 replications. The results suggest that the estimated optimal value is close to the actual one and the estimated CIs show good accuracy. When the estimation sample size increases from 2000 to 3000,  the empirical coverage goes  closer to the nominal one.

\begin{table}
\spacingset{1}
\caption{{{Estimated optimal value and empirical coverage of confidence intervals for the optimal value
}}} 
\begin{center}\small
\tabcolsep 0.15in\renewcommand{\arraystretch}{1.1} \doublerulesep
1.2pt
\begin{tabular}{lcccc}
\hline \hline
($n,N$)&(2000,5000) & (2000, 15,000)&(3000,5000) & (3000, 15,000)
\\\hline
Estimates&\multicolumn{4}{c}{$E(Q^*_{n})$=0.667}  \\
$\mathbb{P}_N\widehat{Q}^*_{n}$ &0.665&0.663&0.664&0.667\\
$\widehat{\sigma}(\mathbb{P}_N\widehat{Q}^*_{n})$ &0.024&0.023&0.020&0.019 \\
\multicolumn{4}{l}{Coverage} \\
90\% CI &0.838&0.828&0.874&0.864 \\
95\% CI&0.898&0.896&0.932&0.928\\
99\% CI &0.958&0.952&0.988&0.990\\
\hline
\end{tabular}
\end{center}\label{CI}
\small NOTE: The estimation sample sizes are $n=$ 2000 and 3000, and the testing sample sizes are~$N=$ 5000 and 15,000. The results are the empirical coverage of 90\%, 95\% and 99\% CI from $500$ replications,   together with the mean of the estimated optimal values, $\mathbb{P}_N(\widehat{Q}^*_{n})$, and the mean of the estimated standard deviations,~$\widehat{\sigma}(\mathbb{P}_N\widehat{Q}^*_{n})$.
\end{table}

Finally, we construct confidence intervals for the difference between the optimal value and the value of a pre-specified decision rule $\pi(\cdot)$. We build confidence intervals according to Proposition~\ref{CLT_Value_Dif} and use the feasible standard deviation estimator in \eqref{hat_sig_2}.
We calculate the empirical coverage from 500 replications. The results are  in Table~\ref{TEST_Dif}. The results show that the estimated value difference is close to the true one for different  estimation sample sizes and testing sample sizes. Moreover,  the estimated CIs exhibit good accuracy. When the estimation sample size increases from 2000 to 3000, the empirical coverage of the confidence interval goes closer to the nominal level. 

\begin{table}
\spacingset{1}
\caption{{{Estimated value differences and the empirical coverage of confidence intervals for the value difference between the optimal decision and pre-specified decision rules.  }}}
\begin{center}\small
\tabcolsep 0.1in\renewcommand{\arraystretch}{1.1} \doublerulesep
1.2pt
\begin{tabular}{lcccc}
\hline \hline
($n,N$)&(2000,5000) & (2000, 15,000)&(3000,5000) & (3000, 15,000)\\\hline
 Estimates&\multicolumn{4}{c}{$E(Q^*_{n})-E(Q_{n,\pi_1})=1.037$} \\
$\mathbb{P}_N\widehat{Q}^*_{n}-\mathbb{P}_N\widehat{Q}_{n,\pi_1}$ &1.034&1.030& 1.039&1.040 \\
$\widehat{\sigma}(\mathbb{P}_N\widehat{Q}^*_{n}-\mathbb{P}_N\widehat{Q}_{n,\pi_1})$ &0.032&0.030&0.026&0.025 \\
\multicolumn{4}{l}{Coverage}\\
90\% CI &0.830&0.816&0.856&0.844\\
95\% CI&0.896&0.888&0.914&0.902\\
99\% CI&0.958&0.954&0.978&0.978\\\\
 Estimates&\multicolumn{4}{c}{$E(Q^*_{n})-E(Q_{n,\pi_2})=1.734$} \\
$\mathbb{P}_N\widehat{Q}^*_{n}-\mathbb{P}_N\widehat{Q}_{n,\pi_2}$ &1.736&1.732&1.730&1.733  \\
$\widehat{\sigma}(\mathbb{P}_N\widehat{Q}^*_{n}-\mathbb{P}_N\widehat{Q}_{n,\pi_2})$ &0.036&0.036&0.030&0.029 \\
\multicolumn{4}{l}{Coverage}\\
90\% CI &0.846&0.846&0.866&0.862 \\
95\% CI&0.902&0.908&0.928&0.920\\
99\% CI&0.962&0.958&0.988&0.988 \\
\hline
\end{tabular}
\end{center}\label{TEST_Dif}
\small NOTE: The estimation sample sizes are $n=$ 2000 and 3000, and the testing sample sizes are~$N=$ 5000 and 15,000. The results are the empirical coverage of 90\%, 95\% and~99\% CI from 500 replications,  together with the mean of the estimated value  difference $\mathbb{P}_N\widehat{Q}^*_{n}-\mathbb{P}_N\widehat{Q}_{n,\pi}$ and the mean of the estimated standard deviation $\widehat{\sigma}(\mathbb{P}_N\widehat{Q}^*_{n}-\mathbb{P}_N\widehat{Q}_{n,\pi})$.
\end{table}

More simulation examples under different simulation settings  with various nonsparsity degree $s_n$ and various choices of $L_n$ are in Appendix G of the supplementary materials. The results show robustness in the performance of our estimator under various settings.

In summary, the simulation results show that our proposed methodology gives rise to better model estimation  accuracy than the standard penalized regression when sparsity occurs to the linear projection of the coefficients. Our approach also allows for valid inferences of the optimal value and value differences using the feasible variance estimator, which performs  well in terms of empirical coverage.

\section{Empirical Study}\label{Empi}
\subsection{Individualized Asset Allocation Optimization}\label{emp_prob_sett}
We study the asset allocation optimization problem under a consumption-based utility framework.  Specifically, we adopt the additive utility with constant relative risk aversion (CRRA, e.g., \citealt{hall1978stochastic}). The utility has the form~$U(C_{1}, C_{2}, ..., C_{T})=\sum_{t=1}^T\gamma^{t}u(C_{t})$, where $C_{t}$ is the consumption at time~$t$,~$u(C)=C^{1-\rho}/(1-\rho)$, $\rho$ is the risk aversion parameter that reflects the sensitivity of utility to the randomness in the income, and $\gamma$ is the discount rate (set as $0.96$; see, e.g., \citealt{Jonathan2002}). The consumption evolves according to the net income~$I_t$ and wealth $W_{t}$ constraint~\hbox{$I_t-C_{t}=W_{t+1}-W_{t}$}. 
The objective is to  maximize the expected utility by choosing the optimal proportion of the total financial asset that is invested in stocks, i.e., stock ratio,~$A$. The utility optimization problem spans the preceding five years (i.e.,~\hbox{$T=5$}). 

Formally, the optimal stock ratio solves the following optimization problem\footnote{In this exploratory study, we consider a simplified situation where utilities are only determined by the consumption within the next five years. This does not mean, however, that individuals do not care what happens after five years. In fact, in our model, consumption is considered optimal with respect to wealth, in particular,  $C_5$ depends on $W_5$. An individual's preference for a high $C_5$ is typically consistent with a high $W_5$, which means that one does care about having enough savings for five years later. In future works where multi-stage decision-making is studied, this would become even less an issue. }:
$
\pi^*(\mathbf{X})=\mathop{\text{argmax}}_{A\in [0,1]} E\big(U(C_{1}, C_{2}, ...C_{5}|\mathbf{X},A)\big),
$
where $\mathbf{X}$ is the personal characteristics. 
The asset allocation decision $A$ determines the  distribution of income and consequently influences the dynamics of wealth and consumption. We consider  randomness in income from not only the financial returns but also from other sources such as wages and medical expenses, the distributions of which vary from individual to individual. Furthermore, the risk aversion parameter $\gamma$ also differs between individuals and relates to their personal characteristics.  

\subsection{Data}
We use the Health and Retirement Study (HRS) data from 1992 to 2014. HRS is a national-level
longitudinal survey of more than 22,000 U.S. residents over the age of~50.
 We include key variables on households finance (e.g., total financial wealth, value of stocks held and income) and individual characteristics (e.g., age, marital status, education level,  health condition and working status) in our analysis. We also incorporate  the  Consumption and Activities Mail Survey (CAMS) data that contain consumption information, which began in~2002,  complementing the HRS. The CAMS data cover approximately 6,800 people, which is a subset of respondents covered in the HRS data.  The relevant variable in~CAMS that we incorporate in our analysis is the nondurable consumption. The raw HRS and CAMS data are noisy and contain many missing observations. In addition, the observed stock ratios are concentrated around zero. Instead of using the raw data, we use pseudo observations that are randomly generated from models of trajectory paths for consumption and income. We also learn risk aversion parameters by matching observed investment profiles with individual characteristics. Details of the model-based random experiment generation are in Appendix~H of the supplementary materials.

\subsection{Statistical Learning Implementation and  Results}
We conduct the analysis based on three samples:  \emph{estimation sample}, \emph{evaluation sample} and \emph{testing sample}. For the observations that are covered by both CAMS and HRS, we split them into an estimation sample and an evaluation sample. The estimation of the Q-function is performed on the \emph{estimation sample}, which contains 2,000 observations. The model fitting is evaluated on the \emph{evaluation sample}, which includes 1,000 observations.  We put all the~15,000+ observations that are not used for the model estimation in the \emph{testing sample}, which includes those that are not covered by~CAMS. The \emph{testing sample} is used in evaluating the value of strategies.

\subsubsection{Q-function estimation}
   
   We include the following covariates in the Q-function~\eqref{Q_learning_presentation}: gender, education, unhealthiness, working status, marital  status, age, household wealth, income, and an intercept term, thus, $d=9$. We follow the common practice in asset allocation applications and adopt the 0.1 incrementation in stock ratios. The stock ratios are divided into 11 levels~$0, 0.1, ..., 1$ (i.e.,~$L_n=11$).  The total number of coefficients is therefore~$p=9\times 11=99$. All covariates are standardized to have a mean of 0 and standard deviation of 0.1. 
For each individual $i$ in the estimation sample, we randomly assign a stock ratio~$A_i$ (0, 0.1, ..., 1) and perform the model-based random experiments.  We set the utility reward variable $Y_{i}$ to be a {Gaussianized score} based on the rank of the average utility achieved by $A_i$ within all stock ratio levels;  see Appendix H.3.  We estimate the value function  \eqref{Q_learning_presentation} using these~2000 observations in the estimation sample. The details of the personal characteristics included in the model and the estimated model coefficients are in Appendix I of the supplementary materials. 

We then check the goodness of fit of the model on the evaluation sample of size~1000. The evaluation sample is different from the estimation sample but within the subset covered by the~CAMS that has consumption information, and hence the utility scores are obtainable. The  out-of-sample~$R^2$ is~31.4\% (in-sample~$R^2$:~45.0\%), suggesting a good fit.

\subsubsection{Strategy performance}

We evaluate the performance of our estimated  individualized optimal asset allocation on the testing sample ($N\geq$15,000). The testing sample contains the individuals in HRS who are not included in the  estimation sample. Because most observations of the testing sample are not covered by~CAMS, consumption information is scarce. Nevertheless, because our final Q-function does not include the consumption variable, using the estimated Q-function, we can evaluate the performance of various strategies  on the testing sample.

We compare the performance of our individualized optimal asset allocation strategy\footnote{{The proposed strategy is based on the {DROVE} estimator we develope. We also evaluated the strategy based on standard SCAD estimator (std-scad) used in \cite{shi2016robust} by direct discretization to the continuous-action  setting. Compared to the std-scad strategy, our  {DROVE} approach achieves a higher average value (0.619 vs. 0.597), and the difference between the two is statistically significant.  }}
$\widehat{\pi}^*_{n}$ with the following benchmark strategies:
(a) we assign the same fixed (0, 0.1, ..., 1) stock ratio to all individuals, a strategy denoted by $\pi_{j}=j, j=0, 0.1, 0.2, ..., 1$; 
(b) we assign the originally observed stock ratio, which is denoted by $\pi_{obs}$.

 For each strategy, we compute the estimated policy value $\mathbb{P}_N \widehat{Q}_{n,\pi}=\sum_{i=1}^N\widehat{Q}_{n}(\mathcal{X}_i, \pi(\mathcal{X}_i))/N$. 
 We construct CIs for $E(Q^*_{n})-E(Q_n(\mathbf{\mathcal{X}}, \pi))$ according to Proposition~\ref{CLT_Value_Dif}. 
 The results are in Table~\ref{Empi_value}.

The results in Table~\ref{Empi_value} suggest that our individualized optimal asset allocation decisions substantially improve over the benchmark strategies in terms of policy values. Compared to the observed stock ratios, the  individualized strategy improves the average utility reward from 0.216 to 0.619. It also yields an average utility reward that is~$0.316$--$1.708$ greater, relative to the fixed strategies. 
The 95\% CIs of the value differences are away from zero,  suggesting that the individualized optimal strategy achieves a significantly higher value compared with the values of the observed  and fixed strategies.  This result demonstrates the importance of  individualization in asset allocation.

\begin{table}[H]
\spacingset{1}
\caption{{Performance of various strategies. }} 
\begin{center}\small
\tabcolsep 0.15in\renewcommand{\arraystretch}{1.1} \doublerulesep
1.2pt
\begin{tabular}{ccccc}
\hline \hline
Strategy  &$\mathbb{P}_N\widehat{Q}_{n,\pi}$&$\mathbb{P}_N\widehat{Q}^*_{n}-\mathbb{P}_N\widehat{Q}_{n,\pi}$&\multicolumn{2}{c}{ 95\% CI of ($EQ^*_{n}-EQ_{n,\pi}$)}\\\hline
$\pi_{obs}$&0.216&0.403&0.328&0.479\\
$\pi_{0}$&0.303&0.316&0.227&0.406\\
$\pi_{0.1}$&0.238&0.381&0.268&0.493\\
$\pi_{0.2}$&0.244&0.375&0.273&0.477\\
$\pi_{0.3}$&0.272&0.347&0.255&0.439\\
$\pi_{0.4}$&0.267&0.352&0.250&0.454\\
$\pi_{0.5}$&-0.382&1.001&0.857&1.145\\
$\pi_{0.6}$&-0.334&0.953&0.816&1.091\\
$\pi_{0.7}$&-0.362&0.981&0.841&1.121\\
$\pi_{0.8}$&-0.374&0.993&0.850&1.137\\
$\pi_{0.9}$&-1.089&1.708&1.546&1.870\\
$\pi_{1.0}$&-1.079&1.698&1.541&1.855\\
\bf$\hat{\pi}^*_{n}$&\bf 0.619&--&--&--\\\hline
\end{tabular}
\end{center}\label{Empi_value}
\footnotesize NOTE:  This table gives the results of the value comparison among the estimated optimal strategy, ($\widehat{\pi}^*_{n}$), observed strategy, ($\pi_{obs}$), and fixed strategies, ($\pi_{j}=j$, $j=0, 0.1, ...,1$). The values are the estimated value (i.e., the average estimated utility reward) for various strategies, $\mathbb{P}_N\widehat{Q}_{n,\pi}$, the estimated value improvement when using the optimal strategy over the benchmark strategies, $\mathbb{P}_N\widehat{Q}^*_{n}-\mathbb{P}_N\widehat{Q}_{n,\pi}$, and the 95\% CIs of the value difference $EQ^*_{n}-EQ_{n,\pi}$.   The model estimation uses a random estimation sample of 2000. The testing sample comprises the remaining 15,000+ observations in the HRS from distinct individuals. 
\end{table}

A low proportion of households participate in the equity market; few of them hold stocks (e.g., \citealt{hong2004social,campbell2006household}). We also find a low participation in equities. Notably, when adopting the optimal strategy, stock market participation increases from 38\% to 63\%; see Appendix J of the supplementary materials for more details. These results show that our optimal individualized strategy improves the financial well-being of the population and it entails a higher stock market participation on average, which indicates a healthier and more active stock market. 
\section{Conclusion and Discussion}\label{Conc}
We develop a high-dimensional statistical learning methodology for continuous-action decision-making with an important application in individualized asset allocation. We show that our DROVE approach enjoys consistency in the model coefficients estimation. Moreover, our approach achieves valid statistical inference on the optimal value.  Empirically, we apply the proposed methodology to study the individualized asset allocation problem using HRS and CAMS data.  Under a consumption-based utility framework, our individualized optimal asset allocation strategy  substantially improves the financial well-being of the population. The outperformance of our individualized optimal strategy over the fixed stock ratio strategies highlights the benefit of individualization.

The statistical learning framework developed in this paper has broad implications. 
 {Methodologically,  our approach can  be extended to the study of stochastic policies, which can have advantages for problems with partially observed states (e.g., \citealt{singh1994learning}) and be of interest for applications such as mobile health under infinite horizon settings (e.g.,  \citealt{luckett2019estimating,liao2021off})}. Another important direction is to extend our framework from the single period to multiple-period decision-making.  {For multiple-stage problems, one potential approach would be to model the Q-function for each stage using the method developed in this paper and then perform a backward recursive procedure to obtain the optimal dynamic decisions. About estimation and inference of the optimal value, challenges arise in the multi-stage problem given that the estimation errors in the optimal rule and the associated optimal value in the latter stage carry over to the former stages. 
} In terms of application, for example, not limited to the setting exercised in this paper, the framework can be readily applied to achieving other wealth management objectives, such as  post-retirement saving adequacy.  The framework can also be extended to the study of multi-class asset allocation, and to incorporate dynamic prediction models of mean and volatility  of financial returns.

\section*{Supplementary Materials}
The online supplementary materials contain the additional assumptions, proofs of the main results in the article, and additional discussions on the theoretical and numerical results.
\section*{Acknowledgement}
We thank the editor, the associate editor, and two anonymous referees for very constructive comments.
\section*{Funding}
Research is partially supported by startup fund of Hong Kong Polytechnic University,  RGC GRF 15302321, RGC GRF16502118 and T31-604/18N of the HKSAR, and NSF-DMS-1555244. 
\section*{Disclosure statement}
The authors report there are no competing interests to declare. 

\bibliographystyle{apalike}
\bibliography{Draft_PersonalizedInvest_main}

\begin{thebibliography}{}

\bibitem[Arnold and Tibshirani, 2016]{arnold2016efficient}
Arnold, T.~B. and Tibshirani, R.~J. (2016).
\newblock Efficient implementations of the generalized lasso dual path
  algorithm.
\newblock {\em Journal of Computational and Graphical Statistics}, 25(1):1--27.

\bibitem[Athey and Wager, 2019]{athey2019efficient}
Athey, S. and Wager, S. (2019).
\newblock Efficient policy learning.
\newblock {\em arXiv preprint arXiv:1702.02896}.

\bibitem[Banks et~al., 1998]{banks1998there}
Banks, J., Blundell, R., and Tanner, S. (1998).
\newblock Is there a retirement-savings puzzle?
\newblock {\em American Economic Review}, pages 769--788.

\bibitem[Cai et~al., 2020]{cai2020deep}
Cai, H., Shi, C., Song, R., and Lu, W. (2020).
\newblock Deep jump q-evaluation for offline policy evaluation in continuous
  action space.
\newblock {\em arXiv preprint arXiv:2010.15963}.

\bibitem[Campbell, 2006]{campbell2006household}
Campbell, J.~Y. (2006).
\newblock Household finance.
\newblock {\em The Journal of Finance}, 61(4):1553--1604.

\bibitem[Chen et~al., 2016]{chen2016personalized}
Chen, G., Zeng, D., and Kosorok, M.~R. (2016).
\newblock Personalized dose finding using outcome weighted learning.
\newblock {\em Journal of the American Statistical Association},
  111(516):1509--1521.

\bibitem[Chernozhukov et~al., 2017]{chernozhukov2017double}
Chernozhukov, V., Chetverikov, D., Demirer, M., Duflo, E., Hansen, C., and
  Newey, W. (2017).
\newblock Double/debiased/neyman machine learning of treatment effects.
\newblock {\em American Economic Review}, 107(5):261--65.

\bibitem[De~Nardi and Yang, 2014]{de2014bequests}
De~Nardi, M. and Yang, F. (2014).
\newblock Bequests and heterogeneity in retirement wealth.
\newblock {\em European Economic Review}, 72:182--196.

\bibitem[Engen et~al., 1999]{engen1999adequacy}
Engen, E.~M., Gale, W.~G., Uccello, C.~E., Carroll, C.~D., and Laibson, D.~I.
  (1999).
\newblock The adequacy of household saving.
\newblock {\em Brookings Papers on Economic Activity}, 1999(2):65--187.

\bibitem[Fan and Li, 2001]{fan2001variable}
Fan, J. and Li, R. (2001).
\newblock Variable selection via nonconcave penalized likelihood and its oracle
  properties.
\newblock {\em Journal of the American Statistical Association},
  96(456):1348--1360.

\bibitem[Fan and Lv, 2011]{fan2011nonconcave}
Fan, J. and Lv, J. (2011).
\newblock Nonconcave penalized likelihood with np-dimensionality.
\newblock {\em IEEE Transactions on Information Theory}, 57(8):5467--5484.

\bibitem[Fan et~al., 2014]{fan2014strong}
Fan, J., Xue, L., and Zou, H. (2014).
\newblock Strong oracle optimality of folded concave penalized estimation.
\newblock {\em The Annals of Statistics}, 42(3):819.

\bibitem[Gourinchas and Parker, 2002]{Jonathan2002}
Gourinchas, P.-O. and Parker, J.~A. (2002).
\newblock Consumption over the life cycle.
\newblock {\em Econometrica}, 70(1):47--89.

\bibitem[Haider and Stephens~Jr, 2007]{haider2007there}
Haider, S.~J. and Stephens~Jr, M. (2007).
\newblock Is there a retirement-consumption puzzle? evidence using subjective
  retirement expectations.
\newblock {\em The Review of Economics and Statistics}, 89(2):247--264.

\bibitem[Hall, 1978]{hall1978stochastic}
Hall, R.~E. (1978).
\newblock Stochastic implications of the life cycle-permanent income
  hypothesis: theory and evidence.
\newblock {\em Journal of Political Economy}, 86(6):971--987.

\bibitem[Hong et~al., 2004]{hong2004social}
Hong, H., Kubik, J.~D., and Stein, J.~C. (2004).
\newblock Social interaction and stock-market participation.
\newblock {\em The Journal of Finance}, 59(1):137--163.

\bibitem[Hurd and Rohwedder, 2003]{hurd2003retirement}
Hurd, M. and Rohwedder, S. (2003).
\newblock The retirement-consumption puzzle: Anticipated and actual declines in
  spending at retirement.
\newblock Technical report, National Bureau of Economic Research.

\bibitem[Kennedy et~al., 2017]{kennedy2017nonparametric}
Kennedy, E.~H., Ma, Z., McHugh, M.~D., and Small, D.~S. (2017).
\newblock Nonparametric methods for doubly robust estimation of continuous
  treatment effects.
\newblock {\em Journal of the Royal Statistical Society. Series B, Statistical
  Methodology}, 79(4):1229.

\bibitem[Laber and Zhao, 2015]{laber2015tree}
Laber, E. and Zhao, Y. (2015).
\newblock Tree-based methods for individualized treatment regimes.
\newblock {\em Biometrika}, 102(3):501--514.

\bibitem[Liao et~al., 2021]{liao2021off}
Liao, P., Klasnja, P., and Murphy, S. (2021).
\newblock Off-policy estimation of long-term average outcomes with applications
  to mobile health.
\newblock {\em Journal of the American Statistical Association},
  116(533):382--391.

\bibitem[Luckett et~al., 2019]{luckett2019estimating}
Luckett, D.~J., Laber, E.~B., Kahkoska, A.~R., Maahs, D.~M., Mayer-Davis, E.,
  and Kosorok, M.~R. (2019).
\newblock Estimating dynamic treatment regimes in mobile health using
  v-learning.
\newblock {\em Journal of the American Statistical Association}.

\bibitem[Lv and Fan, 2009]{lv2009unified}
Lv, J. and Fan, Y. (2009).
\newblock A unified approach to model selection and sparse recovery using
  regularized least squares.
\newblock {\em The Annals of Statistics}, pages 3498--3528.

\bibitem[Meinshausen and B{\"u}hlmann, 2006]{meinshausen2006high}
Meinshausen, N. and B{\"u}hlmann, P. (2006).
\newblock High-dimensional graphs and variable selection with the lasso.
\newblock {\em The Annals of Statistics}, pages 1436--1462.

\bibitem[Munnell et~al., 2012]{munnell2012national}
Munnell, A.~H., Webb, A., Golub-Sass, F., et~al. (2012).
\newblock The national retirement risk index: An update.
\newblock {\em Center for Retirement Research at Boston College}, 1:719--744.

\bibitem[Murphy, 2003]{murphy2003optimal}
Murphy, S.~A. (2003).
\newblock Optimal dynamic treatment regimes.
\newblock {\em Journal of the Royal Statistical Society: Series B (Statistical
  Methodology)}, 65(2):331--355.

\bibitem[Murphy, 2005]{murphy2005experimental}
Murphy, S.~A. (2005).
\newblock An experimental design for the development of adaptive treatment
  strategies.
\newblock {\em Statistics in Medicine}, 24(10):1455--1481.

\bibitem[Palumbo, 1999]{palumbo1999uncertain}
Palumbo, M.~G. (1999).
\newblock Uncertain medical expenses and precautionary saving near the end of
  the life cycle.
\newblock {\em The Review of Economic Studies}, 66(2):395--421.

\bibitem[Qian and Murphy, 2011]{qian2011performance}
Qian, M. and Murphy, S.~A. (2011).
\newblock Performance guarantees for individualized treatment rules.
\newblock {\em The Annals of Statistics}, 39(2):1180.

\bibitem[Robins, 2004]{robins2004optimal}
Robins, J.~M. (2004).
\newblock Optimal structural nested models for optimal sequential decisions.
\newblock In {\em Proceedings of the Second Seattle Symposium in
  Biostatistics}, pages 189--326. Springer.

\bibitem[Rosen and Wu, 2004]{rosen2004portfolio}
Rosen, H.~S. and Wu, S. (2004).
\newblock Portfolio choice and health status.
\newblock {\em Journal of Financial Economics}, 72(3):457--484.

\bibitem[Rubin, 1974]{rubin1974estimating}
Rubin, D.~B. (1974).
\newblock Estimating causal effects of treatments in randomized and
  nonrandomized studies.
\newblock {\em Journal of Educational Psychology}, 66(5):688.

\bibitem[She, 2010]{she2010sparse}
She, Y. (2010).
\newblock Sparse regression with exact clustering.
\newblock {\em Electronic Journal of Statistics}, 4:1055--1096.

\bibitem[Shi et~al., 2018]{Shi2017}
Shi, C., Fan, A., Song, R., and Lu, W. (2018).
\newblock High-dimensional a-learning for optimal dynamic treatment regimes.
\newblock {\em The Annals of Statistics}, 46(3):925.

\bibitem[Shi et~al., 2016]{shi2016robust}
Shi, C., Song, R., and Lu, W. (2016).
\newblock Robust learning for optimal treatment decision with
  np-dimensionality.
\newblock {\em Electronic Journal of Statistics}, 10:2894.

\bibitem[Singh et~al., 1994]{singh1994learning}
Singh, S.~P., Jaakkola, T., and Jordan, M.~I. (1994).
\newblock Learning without state-estimation in partially observable markovian
  decision processes.
\newblock In {\em Machine Learning Proceedings 1994}, pages 284--292. Elsevier.

\bibitem[Tibshirani, 1996]{tibshirani1996regression}
Tibshirani, R. (1996).
\newblock Regression shrinkage and selection via the lasso.
\newblock {\em Journal of the Royal Statistical Society. Series B
  (Methodological)}, pages 267--288.

\bibitem[Tibshirani et~al., 2005]{tibshirani2005sparsity}
Tibshirani, R., Saunders, M., Rosset, S., Zhu, J., and Knight, K. (2005).
\newblock Sparsity and smoothness via the fused lasso.
\newblock {\em Journal of the Royal Statistical Society: Series B (Statistical
  Methodology)}, 67(1):91--108.

\bibitem[Tibshirani and Taylor, 2011]{tibshirani2011solution}
Tibshirani, R.~J. and Taylor, J. (2011).
\newblock The solution path of the generalized lasso.
\newblock {\em The Annals of Statistics}, 39(3):1335--1371.

\bibitem[Tibshirani and Taylor, 2012]{tibshirani2012degrees}
Tibshirani, R.~J. and Taylor, J. (2012).
\newblock Degrees of freedom in lasso problems.
\newblock {\em The Annals of Statistics}, 40(2):1198--1232.

\bibitem[Wager and Athey, 2018]{wager2018estimation}
Wager, S. and Athey, S. (2018).
\newblock Estimation and inference of heterogeneous treatment effects using
  random forests.
\newblock {\em Journal of the American Statistical Association},
  113(523):1228--1242.

\bibitem[Wang et~al., 2013]{wang2013calibrating}
Wang, L., Kim, Y., and Li, R. (2013).
\newblock Calibrating non-convex penalized regression in ultra-high dimension.
\newblock {\em The Annals of Statistics}, 41(5):2505.

\bibitem[Watkins, 1989]{watkins1989learning}
Watkins, C. (1989).
\newblock {\em Learning from delayed rewards}.
\newblock PhD thesis, King's College, Cambridge.

\bibitem[Zhang et~al., 2010]{zhang2010nearly}
Zhang, C.-H. et~al. (2010).
\newblock Nearly unbiased variable selection under minimax concave penalty.
\newblock {\em The Annals of Statistics}, 38(2):894--942.

\bibitem[Zhang and Huang, 2008]{zhang2008sparsity}
Zhang, C.-H. and Huang, J. (2008).
\newblock The sparsity and bias of the lasso selection in high-dimensional
  linear regression.
\newblock {\em The Annals of Statistics}, pages 1567--1594.

\bibitem[Zhao and Yu, 2006]{zhao2006model}
Zhao, P. and Yu, B. (2006).
\newblock On model selection consistency of lasso.
\newblock {\em Journal of Machine Learning Research}, 7(Nov):2541--2563.

\bibitem[Zhao et~al., 2012]{zhao2012estimating}
Zhao, Y., Zeng, D., Rush, A.~J., and Kosorok, M.~R. (2012).
\newblock Estimating individualized treatment rules using outcome weighted
  learning.
\newblock {\em Journal of the American Statistical Association},
  107(499):1106--1118.

\bibitem[Zhou et~al., 2021]{zhouparsimonious}
Zhou, W., Zhu, R., and Zeng, D. (2021).
\newblock A parsimonious personalized dose-finding model via dimension
  reduction.
\newblock {\em Biometrika}.

\bibitem[Zhu et~al., 2020]{zhu2020kernel}
Zhu, L., Lu, W., Kosorok, M.~R., and Song, R. (2020).
\newblock Kernel assisted learning for personalized dose finding.
\newblock In {\em Proceedings of the 26th ACM SIGKDD International Conference
  on Knowledge Discovery \& Data Mining}, pages 56--65.

\bibitem[Zhu et~al., 2019]{ZZS19}
Zhu, W., Zeng, D., and Song, R. (2019).
\newblock Proper inference for value function in high-dimensional q-learning
  for dynamic treatment regimes.
\newblock {\em Journal of the American Statistical Association},
  114(527):1404--1417.

\bibitem[Zou and Li, 2008]{zou2008one}
Zou, H. and Li, R. (2008).
\newblock One-step sparse estimates in nonconcave penalized likelihood models.
\newblock {\em The Annals of Statistics}, 36(4):1509.

\end{thebibliography}
\end{document}